\documentclass[sigconf,nonacm]{acmart}

\AtBeginDocument{%
  \providecommand\BibTeX{{%
    \normalfont B\kern-0.5em{\scshape i\kern-0.25em b}\kern-0.8em\TeX}}}

\setcopyright{acmcopyright}
\copyrightyear{2021}
\acmYear{2021}

\usepackage{caption}
\usepackage{subcaption} 
\usepackage{xcolor} 
\usepackage{array} 
\usepackage{tikz}
\DeclareRobustCommand*\circled[1]{\tikz[baseline=(char.base)]{
            \node[shape=circle,draw,inner sep=2pt] (char) {#1};}}
\usepackage{graphicx}

\begin{document}

\title[Explaining Prototypes for Interpretable Image Recognition]{This Looks Like That, Because ... Explaining Prototypes for Interpretable Image Recognition}

\author{Meike Nauta}
\email{m.nauta@utwente.nl}
\orcid{0000-0002-0558-3810}
\affiliation{
  \institution{University of Twente}
  \city{Enschede}
  \country{The Netherlands}}

\author{Annemarie Jutte}
\email{a.m.p.jutte@student.utwente.nl}
\email{j.c.provoost@student.utwente.nl}
\authornote{Both authors contributed equally to this work.}
\author{Jesper Provoost}
\authornotemark[1]
\affiliation{
  \institution{University of Twente}
  \city{Enschede}
  \country{The Netherlands}
}

\author{Christin Seifert}
\email{christin.seifert@uni-due.de}
\affiliation{
 \institution{University of Duisburg-Essen}
  \city{Essen}
  \country{Germany}}
\affiliation{
  \institution{University of Twente}
  \city{Enschede}
  \country{The Netherlands}}

\renewcommand{\shortauthors}{Nauta et al.}

\begin{abstract}
Image recognition with prototypes is considered an interpretable alternative for black box deep learning models. 
Classification depends on the extent to which a test image ``looks like'' a prototype.
However, perceptual similarity for humans can be different from the similarity learned by the classification model. Hence, only visualising prototypes can be insufficient for a user to understand what a prototype exactly represents, and why the model considers a prototype and an image to be similar. 
We address this ambiguity and argue that prototypes should be explained. 
We improve interpretability by automatically enhancing visual prototypes with textual quantitative information about visual characteristics deemed important by the classification model. 
Specifically, our method clarifies the meaning of a prototype by quantifying the influence of colour hue, shape, texture, contrast and saturation and can generate both global and local explanations. Because of the generality of our approach, it can improve the interpretability of any similarity-based method for prototypical image recognition.
In our experiments, we apply our method to the existing Prototypical Part Network (ProtoPNet). 
Our analysis confirms that the global explanations are generalisable, and often correspond to the visually perceptible properties of a prototype. Our explanations are especially relevant for prototypes which might have been interpreted incorrectly otherwise. By explaining such ‘misleading’ prototypes, we improve the interpretability and simulatability of a prototype-based classification model.
We also use our method to check whether visually similar prototypes have similar explanations, and are able to discover redundancy. Code is available at \url{https://github.com/M-Nauta/Explaining_Prototypes}.
\end{abstract}

%
%

%

\begin{teaserfigure}
  \includegraphics[width=\textwidth]{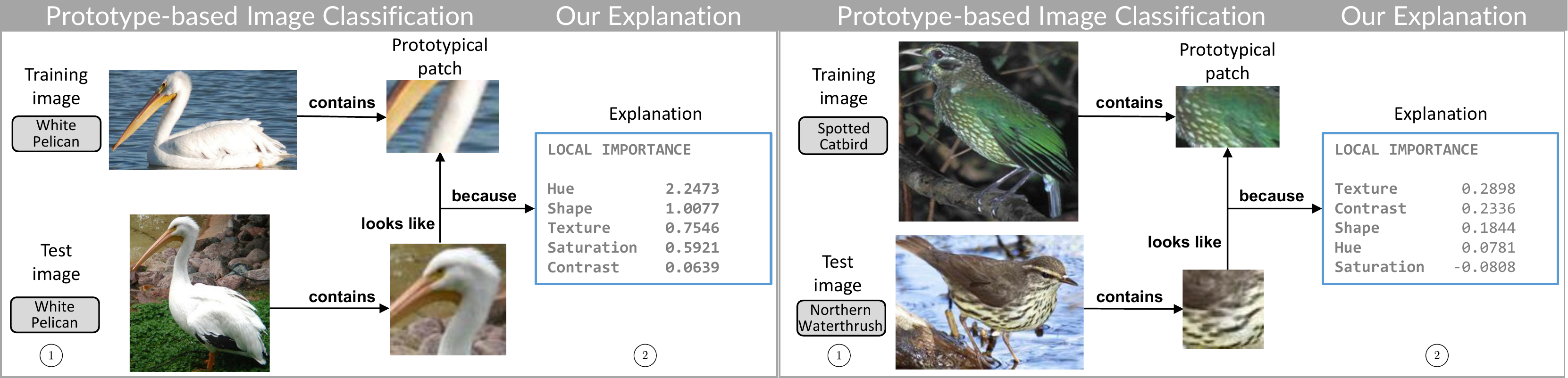}
  \caption{\circled{\small \textmd{1}} Prototype-based image classification (e.g.~ProtoPNet~\cite{chen2019looks}). \circled{\small \textmd{2}} Our contribution: Quantifying the importance of visual characteristics to explain why the classification model deemed an image patch and prototype similar. Left: Logical explanation for the clear similarity between the patches. Right: a `misleading' prototype: humans might expect these patches to be dissimilar, but our method explains that the classification model considers these patches alike because of similar texture.
}
  \label{fig:teaser}
\end{teaserfigure}

\maketitle

\section{Introduction}
Convolutional Neural Networks (CNNs) \cite{lecun1989backpropagation} are the de-facto standard for object detection due to their impressive performance in numerous automated image classification tasks \cite{krizhevsky2012imagenet, simonyan2014very, he2016deep}.
However, the black box nature of neural networks prevents a human to assess the model's decision making process, which is especially problematic in domains with high stakes decisions~\cite{rudin2019stop}. Following this demand on understanding automated decision making, explainable Artificial Intelligence (XAI) has been actively researched ~\cite{guidotti2018survey, adadi2018peeking, arrieta2020explainable}.
\emph{Post-hoc} explanation methods learn a second, transparent model to approximate the first black box model~\cite{guidotti2018survey}, but these reverse-engineering approaches are not guaranteed to show the \emph{actual} reasoning of the black box model~\cite{rudin2019stop}. 
\emph{Intrinsically} interpretable models on the other hand, are faithful by design and allow \emph{simulatability}: a user should be able to reproduce the model's decision making process based on the input data together with the explanations of the interpretable model and come to the same prediction~\cite{chakraborty2017interpretability,Lipton2018mythos}.
One type of such models is prototypical learning~\cite{biehl2016prototype}, which has a transparent, built-in case-based decision making process. We focus on the problem of supervised image recognition where a machine learning model should label an image. Prototypes in this context are usually `nearest neighbours', i.e., images from the training set that look similar to the image being classified ~\cite{bien_prototypes, protoattend, plesse_prototypes, li_prototypes}. The similarity between a prototype and an image is often measured in latent space, learned by the neural network, where images from the same class are close and dissimilar images are far apart with respect to a certain distance or similarity metric.
Recently, the Prototypical Part Network (ProtoPNet)~\cite{chen2019looks} and ProtoTree~\cite{nauta2020neural} were introduced which use prototypical \emph{parts} and identify similar patches in an image. The classification depends on the extent to which \emph{this} part of the image ``looks like'' \emph{that} prototypical part, measured by a similarity score. An example of this reasoning is shown in Fig.~\ref{fig:teaser} (Sect.~\ref{sec:protopnet} discusses ProtoPNet in more detail). 

\textbf{Prototype Ambiguity} In this paper, we address the ambiguity that prototypes can have and present a method to \emph{explain prototypes}. 
Consider the left part in Figure~\ref{fig:teaser}, showing a prototypical patch (`prototype') of a white pelican. Although the similarity between this prototype and the patch in the test image is not surprising, it is unclear what this prototype exactly represents. Is the prototype looking for a white neck, an orange-coloured beak or is the shape of the beak specifically important? The similarity score between the two patches assigned by the model depends on its classification strategy, and hence its learned latent space. 
Explanations are especially needed when similarity is not so obvious. When seeing the two patches in the right part of Figure~\ref{fig:teaser}, 
a human might argue that these patches are dissimilar because of the colour differences. The classification model however assigns these patches a high similarity score, and thus considers them alike, even though the test image is from a different class than the prototype. This shows that a human and the CNN might have different reasoning processes, despite using the same prototypes. 
The classification strategy of a neural network, dependent on the learned latent space, determines the reason for considering two patches as being similar or different. It has been shown that CNNs trained on ImageNet are strongly biased towards recognizing texture~\cite{biased_texture}, although other work shows that CNNs can be biased towards shape~\cite{shapebias} or colour~\cite{colorbias}.
Perceptual similarity for humans however is biased towards shape~\cite{landau1988importance,OpdeBeeck10111}, but also based on e.g. colour, size, semantic similarity, culture and complexity~\cite{segall1966influence,rossion_colour,king2019similarity}. It is also questionable whether humans and CNNs will ever follow the exact same similarity reasoning, since Rosenfeld et al. found that neural networks fall short on predicting human similarity perception~\cite{Rosenfeld_2018_CVPR_Workshops}.
Since a user is not aware of the underlying classification strategy of the trained CNN and might also be unaware of personal biases, only visualising prototypes is insufficient for understanding 
what a prototype exactly represents, and why a prototype and an image are considered similar. 
This issue may also arise with other explainability methods that show or highlight image parts, such as attention mechanisms~\cite{Fu_2017_CVPR,Zheng_2017_ICCV,attention_survey}, components~\cite{components_saralajew} and other part-based models e.g.~\cite{partbased_rcnns_zhang, Zheng_2019_CVPR,zheng_learning_rich}.
Including our explanations can help users to increase the simulatability~\cite{Lipton2018mythos} and general understanding of the model.

\textbf{Contribution}
We improve the interpretability of a prototype-based CNN by automatically enhancing prototypes with extra textual quantitative information about visual characteristics used by the model. Specifically, we present a methodology to quantify the influence of colour hue, saturation, shape, texture, and contrast in a prototype. This clarifies what the model pays attention to and why a model considers two images to be similar. Hence, we disentangle localisation and explanation. Our method can extend any prototype-based model for image recognition, such as ProtoPNet~\cite{chen2019looks} and ProtoTree~\cite{nauta2020neural}. In this paper, we show its applicability for the prototypical parts of ProtoPNet. 
For example, again considering the left part of Figure~\ref{fig:teaser}, our explanation shows that ProtoPNet considers the prototype and patch from the test image to be similar because of the similar colour hue and shape of the beak in the test image.
Our method is especially useful when similarity is not so obvious. It can explain potentially misleading prototypes such as the right prototype in Fig.~\ref{fig:teaser}. Whereas a human might look for something green, our explanation reveals that ProtoPNet considers these two patches similar because of texture, contrast and shape. The similarity is thus because of the dotted pattern and colour hue was not important. This explanation seems reasonable given that the prototype is a patch from the class “Spotted Catbird”.

Our method automatically modifies images to change their hue, shape, texture, contrast or saturation. We forward both the original image and the modified image through ProtoPNet and analyse ProtoPNet's similarity scores. Specifically, the similarity score of ProtoPNet between the prototype and the original image is compared with the similarity score of the prototype and a modified image. The intuition is that a visual characteristic is considered \emph{unimportant} by the classification model when the difference between these two similarity scores is small (and will therefore get a low importance score), and is deemed \emph{important} when the similarity scores differ sufficiently. For example, a blue bird is changed to a purple bird by changing the hue of the image. If hue would have been important for the specific prototype, it would be expected that ProtoPNet assignes a low similarity between the prototype and the purple bird, whereas the similarity with the blue bird was high. As shown in Figure~\ref{fig:teaser}, the prototypes can subsequently be explained by quantifying the importance of visual characteristics. 

The following section summarizes the ProtoPNet model~\cite{chen2019looks}. Section~\ref{sec:methodology} presents our methodology for quantitatively explaining prototypes w.r.t. visual image characteristics. The experimental setup is described in Section~\ref{sec:experimental_setup}, after which results are presented and discussed in Section~\ref{sec:results_discussion}.

\section{Prototypical Part Network}
\label{sec:protopnet}
We apply the methodology presented in this paper to ProtoPNet, the Prototypical Part Network from Chen et al.~\cite{chen2019looks} that follows the ``\textit{this} looks like \textit{that}" reasoning. Prototypical parts learned by ProtoPNet are subsequently explained by our method. 
Key for presenting our explanation methodology is having a global understanding of the workings of ProtoPNet. We refer to the original work by Chen et al.~\cite{chen2019looks} for more specific details on ProtoPNet's training and visualisation process.

\begin{figure}
\centering
\includegraphics[width=0.45\textwidth]{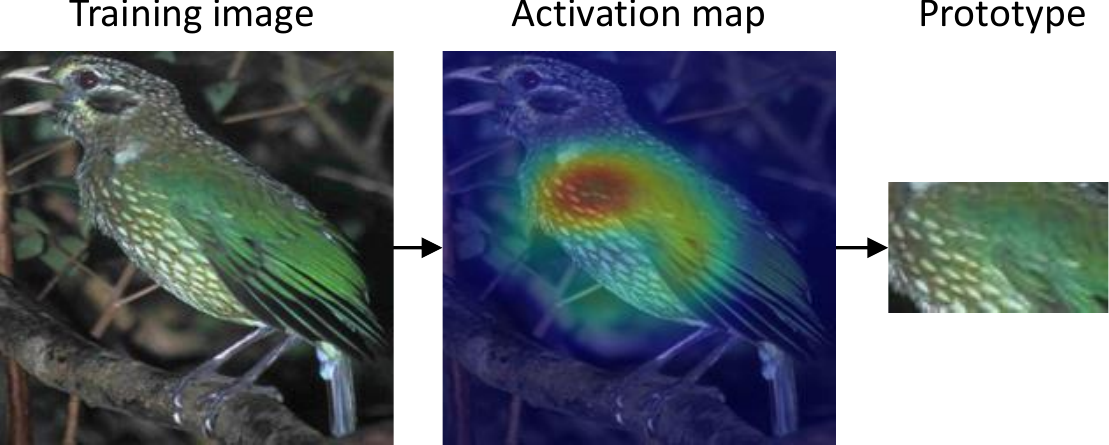}
\caption{A prototype learned by ProtoPNet is the nearest patch of a training image.}
\label{fig:protopnet_process}
\end{figure}

The ProtoPNet architecture consists of a standard CNN (e.g.~ResNet), followed by a prototype layer and a fully-connected layer. The prototype layer consists of a pre-determined number of class-specific prototypes. The implementation of Chen et al.~\cite{chen2019looks} learns 10 prototypes per class. The fully-connected layer learns a weight for each prototype. 
During training, prototypes are vectors in latent space that should learn discriminative, prototypical parts of a class. An input image is forwarded through the CNN, after which the prototype layer compares the resulting latent embedding with the prototype. 
A kernel slides over the latent image and at each location, the squared Euclidean distance between the latent prototype vector and a patch in the latent image is calculated. 
This creates an activation map, containing the distance to the prototype at each location in the latent image. To ensure that the prototype can be visualised, the training procedure of ProtoPNet requires that each prototype is \emph{identical} to some latent training patch. ProtoPNet iterates through all training images and selects the image with the smallest distance to the latent prototype. The corresponding activation map can be upsampled to the size of the original image and visualised as a heatmap (see Figure \ref{fig:protopnet_process}). The prototype can now be visualised as the most similar patch in training images in input space. 

After training, ProtoPNet classifies a test image $k$ by calculating the similarity between a prototype and image $k$.
The distance $d_{j,k}$ between the nearest patch in latent image $k$ to the $j$-th prototype is converted to a similarity score:
\begin{equation}
    g_{j, k} = \log \left(\frac{d_{j,k} + 1}{d_{j,k} + \epsilon}\right),
\end{equation}
where $\epsilon$ is an arbitrarily small positive quantity to prevent zero division.
To classify this image, the similarity scores of the image and each prototype are weighted by the fully-connected layer and summed per class, resulting in a final score for an image belonging to each class. Applying a softmax yields the predicted probability that a given image belongs to a certain class.
The left part of Figure~\ref{fig:protopnet_reasoning+local} shows an illustration of this reasoning process. 

\section{Methodology}
\label{sec:methodology}
In order to obtain importance scores for visual characteristics of prototypes, images are automatically modified. The characteristics in focus are contrast, colour hue and saturation, shape, and texture (cf. Section~\ref{sec:motivation}). Our approach for automatically modifying these characteristics is described in Section~\ref{sec:mods}. Section~\ref{sec:proto_anno} presents a methodology to explain prototypes by quantifying the importance of a visual characteristic. 

\subsection{Important Visual Characteristics}
\label{sec:motivation}
The perceptual and cognitive processing in the human visual system is influenced by various features. To determine which image modifications we need to effectively explain prototypes, we discuss important visual characteristics for the human perceptual system.

The data visualisation domain has a ranking of channels to control the appearance of so-called \emph{marks}~\cite{munzner_visualization_2015}. A `mark' is a basic graphical element in an image, such as a black triangle or moving red dot. Important visual channels for marks are position, size, angle, spatial region, colour hue, colour luminance, colour saturation, curvature, motion and shape~\cite{munzner_visualization_2015}. For static 2-dimensional natural images, motion is not applicable and we consider curvature related to shape. Furthermore, it is not necessary to modify the size, position, angle or spatial region of objects in images, since CNNs with pooling, possibly combined with suitable data augmentation, are invariant to these characteristics~\cite{Goodfellow-et-al-2016, SimonyanZ14a}.
Moreover, research in neuroscience shows that the human eye can recognise objects independent of ambient light level during the day~\cite{striedter2016neurobiology}, whereas contrast (spatial variation in luminance) is needed for edge detection and delineation of objects~\cite{munzner_visualization_2015}. The human visual system is thus more sensitive to contrast than absolute luminance~\cite{striedter2016neurobiology}. We therefore will not modify the absolute luminance, but the contrast in an image. Thus, the visual characteristics from the data visualisation domain that we deem important for explaining a prototype are \textbf{hue}, \textbf{contrast}, \textbf{saturation} and \textbf{shape}. 

The channels for marks mentioned in the previous paragraph are however too simplistic, because they do not include the texture or material of an object. Research in neuroscience also emphasises the importance of texture for classifying objects in the natural world~\cite{pratesi_texture_color_form,kourtzi2000cortical}.
Related to this, Bau et al.~\cite{Bau_2017_CVPR} disentangled visual representations by layers in a CNN and found that self-supervised models, especially in the earlier layers of the network, learn many texture detectors. 
We therefore also include \textbf{texture} as an important visual characteristic.

\subsection{Image Modifications}
\label{sec:mods}
\begin{figure}
\begin{subfigure}{.32\linewidth}
\centering
\includegraphics[width=0.95\linewidth]{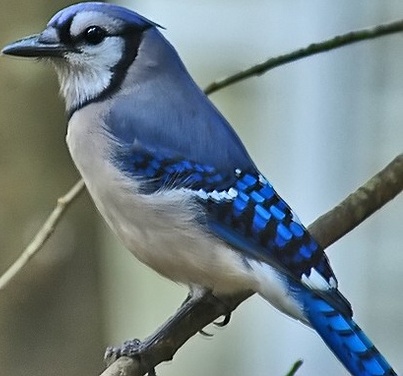}
\caption*{Original}
\end{subfigure}%
\begin{subfigure}{.32\linewidth}
\centering
\includegraphics[width=0.95\linewidth]{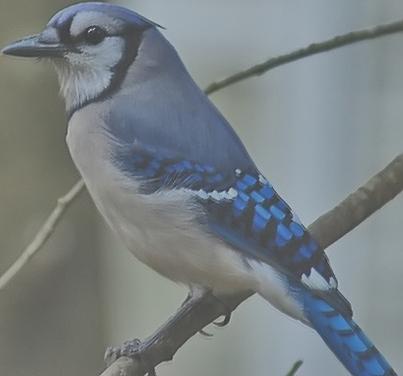}
\caption*{Contrast}
\end{subfigure}%
\begin{subfigure}{.32\linewidth}
\centering
\includegraphics[width=0.95\linewidth]{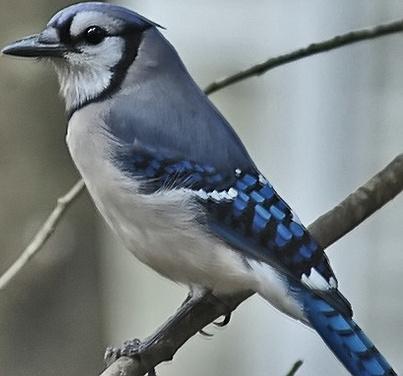}
\caption*{Saturation}
\end{subfigure}
\vspace{3mm}
\begin{subfigure}{.32\linewidth}
\centering
\includegraphics[width=0.95\linewidth]{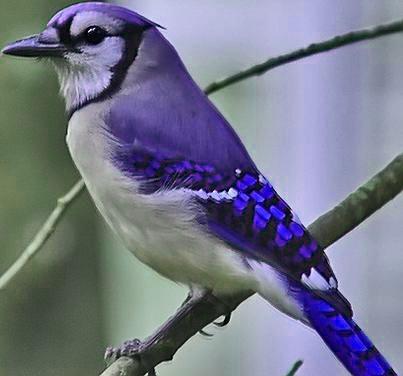}
\caption*{Hue}
\end{subfigure}%
\begin{subfigure}{.32\linewidth}
\centering
\includegraphics[width=0.95\linewidth]{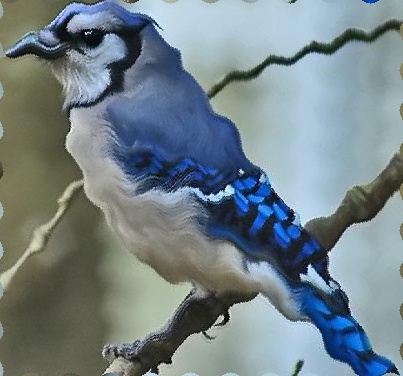}
\caption*{Shape}
\end{subfigure}%
\begin{subfigure}{.32\linewidth}
\centering
\includegraphics[width=0.95\linewidth]{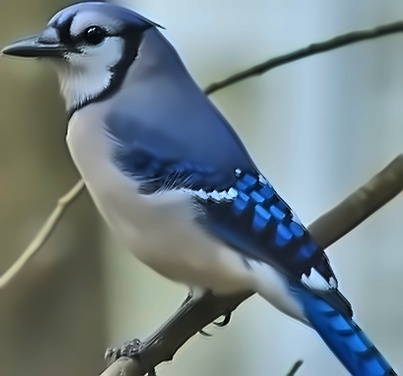}
\caption*{Texture}
\end{subfigure}
\caption{Image modifications for corresponding visual characteristics applied to an exemplar image.}
\label{fig:mods}
\end{figure}

For each of the visual characteristics, an image set is created. Each of these sets contains modified images, which are designed to be harder to classify based on the respective characteristic. For example, we generate a set of low-contrast images, such that contrast information can not (or hardly) be used by the model. Using these modified images, the importance of a characteristic for a specific prototype can be determined by comparing the differences between the prototype-image similarity of the original and modified image. 

To create the modified images, we apply image transformations to reduce the intensity of each characteristic, i.e., we create images with reduced contrast, saturation, hue, shape and texture. Figure~\ref{fig:mods} shows an example image and its modified versions. We opt for automated image modifications instead of manual modifications used for experiments in psychology research (e.g.~\cite{pmlr-v70-ritter17a}), to be able to create a large number of modified images efficiently. 

To create low \textbf{contrast} images, the original image is blended with the mean of its grayscale version. More concretely,  we first create a grayscale version of the image and calculate its mean value. We then generate the modified image by pixel-wise averaging each channel (RGB) of the original image with the mean grayscale value.
Similarly, the low-\textbf{saturation} image is created by averaging the original image with its grayscale counterpart.
To generate an image with different \textbf{colour hues}, the RGB image is converted to the HSV colour space after which the H-dimension is modified for each pixel.
In order to modify \textbf{shapes} in an image, we add a linear displacement by warping the image. Specifically, we shift pixels according to a sine wave in both the horizontal and vertical direction. 
To modify \textbf{texture}, we apply a non-local means denoising technique~\cite{ipol.2011.bcm_nlm}. This method removes small quantities of noise, and can therefore be used to blur the sophisticated texture of a bird while preserving its overall shape. Implementation details are presented in Section~\ref{sec:mod_impl}.

\subsection{Importance Scores for Image Characteristics}
\label{sec:proto_anno}
We evaluate the importance of visual characteristics by calculating a local and a global importance score. 
The \textbf{local} score measures the importance of the visual characteristics for a single image, and is therefore applied on previously unseen images, i.e., any image in the test dataset $S_\text{test}$.
The \textbf{global} score measures the importance of visual characteristics for one prototype in general, and is independent of a specific input image. The global score is obtained from the training dataset $S_\text{train}$.

Let $i \in \{\mathrm{contrast, saturation, hue, shape, texture}\}$ denote the type of modification, $j \in \{1, 2, ..., n\}$ the prototype index and $k$ the image. 
Furthermore, as introduced in Section~\ref{sec:protopnet}, let the similarity of the original image and the prototype be denoted as $g$, and the similarity of the modified image and the prototype be denoted as $\hat{g}$. 
Then the local importance score $\phi^{i, j, k}_\text{local}$ of characteristic $i$ for test image $k \in S_\text{test}$ on the $j$-th prototype is the difference in similarity scores:
\begin{equation}\label{eq:local_score}
    \phi^{i, j, k}_\text{local} = g_{j, k}-\hat{g}_{i, j, k}.
\end{equation}
For the calculation, we fix the patch location such that the part of image $k$ compared with prototype $j$ is the same for both the original and modified image.

Whereas the local importance score indicates to what extent a visual characteristic influences the similarity score given by the prototype-based model \emph{for a single image}, the \emph{global} importance score gives a general impression of the importance of visual characteristics in a prototype. The global score of characteristic $i$ on the $j$th prototype can be calculated by taking all training images into account. A naive approach would be to average over the local scores of all training images. However, prototype $j$ might not be present in all images and modifying those images will therefore not (or barely) influence the resulting similarity score. For example, if prototype $j$ represents a specific beak which is not present in original image $k$, ProtoPNet will give a low similarity score $g_{j, k}$. Since this prototype will also be absent in the modified image, the difference between the similarity scores, Eq.~\ref{eq:local_score}, is near zero. This result could indicate that a certain characteristic is not important, although the result is actually indicating that the prototype was simply not present. Therefore, we create a more informative global importance score by calculating a weighted arithmetic mean by weighing the local scores of all images in $S_\text{train}$ by their similarity score with the prototype:
\begin{equation}
    \phi^{i, j}_\text{global} = \frac{\sum_{k = 1}^{\left|S_\text{train}\right|}\phi^{i, j, k}_\text{local} \cdot g_{j, k}}{\sum_{k = 1}^{\left|S_\text{train}\right|} g_{j, k}}.
\end{equation}
Hence, if unmodified image $k$ gets a low similarity score with prototype $j$, it will get a low weight for the global importance calculation. In contrast, if prototype $j$ is clearly present in image $k$, ProtoPNet will assign a high similarity score and hence $k$ gets a high weight.

These importance scores can be used to create \emph{global} explanations that explain a prototype, and \emph{local} explanations that explain the similarity score between a given image and a prototype. 
The global explanation for the $j$-th prototype lists for each visual characteristic $i$ its importance by showing the importance scores $\phi^{i, j}_\text{global}$. This explanation is thus input independent and can be created before applying the prototype model to unseen images. 
The local explanation is of use during testing and explains a single prediction.

\section{Experimental Setup}
\label{sec:experimental_setup}
To evaluate our method for explaining prototypes, we first train a ProtoPNet~\cite{chen2019looks} that results in an interpretable predictive model with prototypical parts for fine-grained image recognition. We apply our method to the resulting prototypes for generating global and local explanations. Section~\ref{sec:dataset} discusses the dataset, corresponding data augmentation, and hyperparameters for training ProtoPNet. Section~\ref{sec:mod_impl} presents the implementation details for our image modifications. The design of our experiments to evaluate our explanations in presented in Section~\ref{sec:experimental_design}. 

\begin{figure*}
\centering
\includegraphics[width=0.85\textwidth]{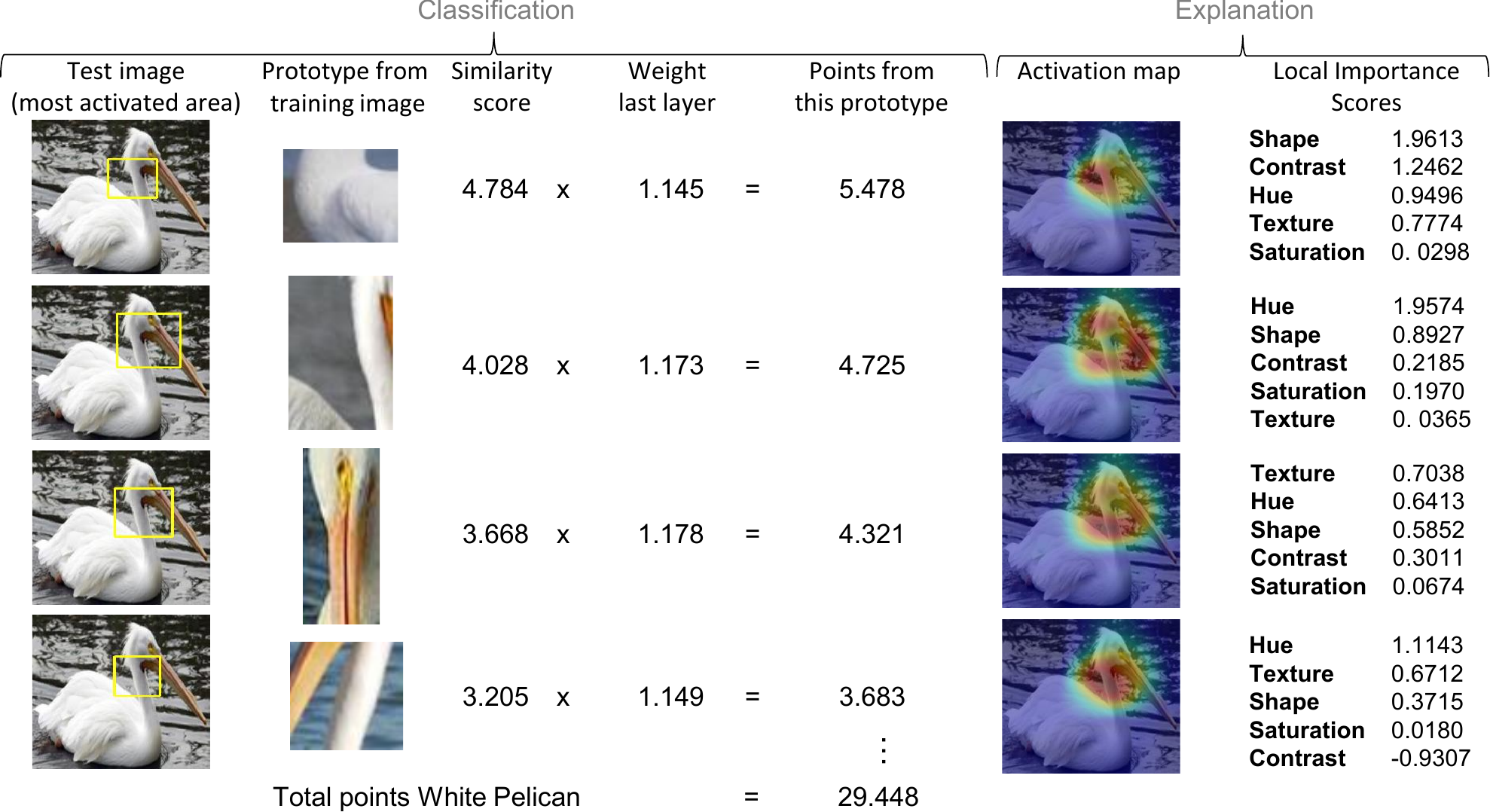}
\caption{Left: ProtoPNet's reasoning with a subset of all prototypes of the White Pelican class. To classify a test image, ProtoPNet compares the class-specific prototypes of each class with the test image to calculate the total number of points for this class. An image is classified as the class with the most points. Right: The activation maps produced by ProtoPNet and our corresponding local explanations that explain which characteristics were important for a similarity score.}
\label{fig:protopnet_reasoning+local}
\end{figure*}

\subsection{Dataset and Parameters}
\label{sec:dataset}
For our experiments, we use the Caltech-UCSD Birds dataset~\cite{WelinderEtal2010}, a dataset for bird species identification which was also used by Chen et al.~\cite{chen2019looks} for training their ProtoPNet. It contains 200 different classes with approximately 60 images per class.
The dataset provides a train-test split, leading to $S_\text{train}$ with 5994 images and $S_\text{test}$ with 5794 images.
For training ProtoPNet, we apply the same data processing techniques as the original work~\cite{chen2019looks}.
We cropped the images according to the bounding boxes provided with the dataset and apply data augmentation on $S_\text{train}$ as described in ProtoPNet's supplementary material~\cite{chen2019looks}. All images are resized to $224 \times 224$.

To train ProtoPNet, we use the code provided by the authors\footnote{\url{https://github.com/cfchen-duke/ProtoPNet}, accessed November, 2020}. We opted for DenseNet-121~\cite{huang2017densely} as pre-trained network for the initial layers of ProtoPNet, as this was reported to be the best-performing network on the Caltech-UCSD dataset~\cite{chen2019looks}. The DenseNet-121 network has been pre-trained on ImageNet~\cite{deng2009imagenet}.\footnote{We use the same methodology as ProtoPNet~\cite{chen2019looks}  in order to reproduce their results, although it is known that there is some overlap between Caltech-USCD and ImageNet.} 
When forwarding the resized images through DenseNet, the input image dimensions, $H_\text{in} = W_\text{in} = 224$ and $D_\text{in} = 3$, are transformed to the output dimensions $H = 7$,  $W = 7$ and $D = 128$. Depth $D$ is a hyperparameter in ProtoPNet determining the number of channels for the network output and the prototypes, and is set to $128$ as in ProtoPNet~\cite{chen2019looks}.
As in the paper by Chen et al.~\cite{chen2019looks} we use 10 prototypes per class, leading to 2,000 prototypes in total. All other training parameters are also replicated from the implementation by Chen et al.~\cite{chen2019looks}.

\subsection{Modification Implementation}
\label{sec:mod_impl}
When implementing the image modifications as described in Section~\ref{sec:mods}, we aim for a similar modification `strength' for all characteristics in order to compare importance scores. Furthermore, the image modifications should be modest, since too extreme modifications can lead to out-of-distribution images that result in erratic behaviour of the underlying neural network of ProtoPNet. 
A similar modification degree depends on how ProtoPNet perceives the images. Therefore, we find a suitable modification degree by forwarding both the unmodified image and the modified image through the underlying CNN and compare their L1-norm distance in latent space. We tune the modifications parameters such that the mean distance between the unmodified latent training images and the latent modified images is exactly $0.0002$ for all characteristics. This value is experimentally chosen such that it results in modifications that are clearly distinguishable for the human eye, while still being perceived by ProtoPNet as being close to the original images. 
For the colour modifications (contrast, saturation and hue), we use PyTorch's image transformations\footnote{\url{https://pytorch.org/docs/stable/torchvision/}, accessed June 2020}. More specifically, we use the \texttt{ColorJitter} function where we set the contrast value to $0.45$, saturation to $0.7$ and hue to $0.1$ for the respective modifications. The shape modification is manually implemented in Python. The texture modification is implemented with the Non-local Means Denoising algorithm~\cite{ipol.2011.bcm_nlm} for coloured images in OpenCV\footnote{\url{https://docs.opencv.org/3.4/d1/d79/group__photo__denoise.html\#ga03aa4189fc3e31dafd638d90de335617}, accessed June 2020}.
The filter strength of the denoising algorithm is set to 4 in order to get the correct mean latent distance. 

\subsection{Considerations for Evaluation}
\label{sec:experimental_design}
We would like to emphasise that we do not want to explain human perception, but the perception of the prototype-based model. Hence, we cannot ask users what they deem important, since we aim to explain the model's reasoning. Also, we cannot construct a ground-truth since we are opening up a ``black-box'' for which no ground-truth is available. 
However, we can still do a quantitative analysis to evaluate the generalizability and robustness of the explanations. If generalised well, one would expect that global importance scores are similar when computed for different image sets. Therefore, we compare the global scores for the training set with global explanations computed from the test set. We also evaluate the distribution of global importance scores to get more insight in the general model reasoning.
Additionally, we qualitatively analyse a varied selection of local explanations (Sect.~\ref{sec:results_local_explanations}) and global explanations (Sect.~\ref{sec:results_global_explanations}). We especially analyse the effectiveness of our explanations for `misleading' prototypes, where our explanations could be different from what a user might expect. Section~\ref{sec:redundant_prototypes} analyses the redundancy of prototypes to answer the question: do visually similar prototypes focus on similar characteristics or do they complement each other?  

\section{Results and Discussion}
\label{sec:results_discussion}
ProtoPNet is trained for 30 epochs, reaching a test accuracy of 78.3\%. Having applied the same data and training process as the original work~\cite{chen2019looks}, we do not know why our accuracy is lower than the accuracy reported in the original work (80.2\%). 
However, the aim of this paper is not to train the best ProtoPNet, but to find a reasonable well-performing model in order to explain its prototypes. 

Figure~\ref{fig:protopnet_reasoning+local} (left) shows a selection of prototypical patches (`prototypes') learned by ProtoPNet. ProtoPNet measures the similarity between a prototype and patches in a given test image. The resulting similarity scores are multiplied with learned weights resulting in a final score per class.
During training, softmax is applied to get a soft prediction. For testing, an image is classified as the class with the highest score (i.e.~most points).  

\subsection{Analysing our Local Explanations}
\label{sec:results_local_explanations}
Figure~\ref{fig:protopnet_reasoning+local} (right) shows how our local explanations complements the prototypical reasoning by explaining which visual characteristics were important for ProtoPNet's similarity score between a prototype and a specific image. We show the activation map as implemented by Chen et al. \cite{chen2019looks} and list the local importances for each visual characteristic. The importances identified by our local explanations for the test image shown in Figure~\ref{fig:protopnet_reasoning+local} seems reasonable given the typical white colour of the pelican and its long neck. Furthermore, our explanations enable a user to understand why ProtoPNet gave a high similarity score to a prototype and a patch in the test image. Whereas the prototypes of the white pelican all look very similar to a human, our local importance scores explain the differences between the prototypes. The topmost prototype in Fig.~\ref{fig:protopnet_reasoning+local} mostly focuses on shape and contrast, the second and fourth prototype deem hue important and the third prototype focuses on texture. Although prototypes might look similar to the human eye, our explanations can differentiate between prototypes and estimate the prototype's purpose.

\begin{figure}
\centering
\begin{subfigure}[b]{0.99\linewidth}
\includegraphics[width=\linewidth]{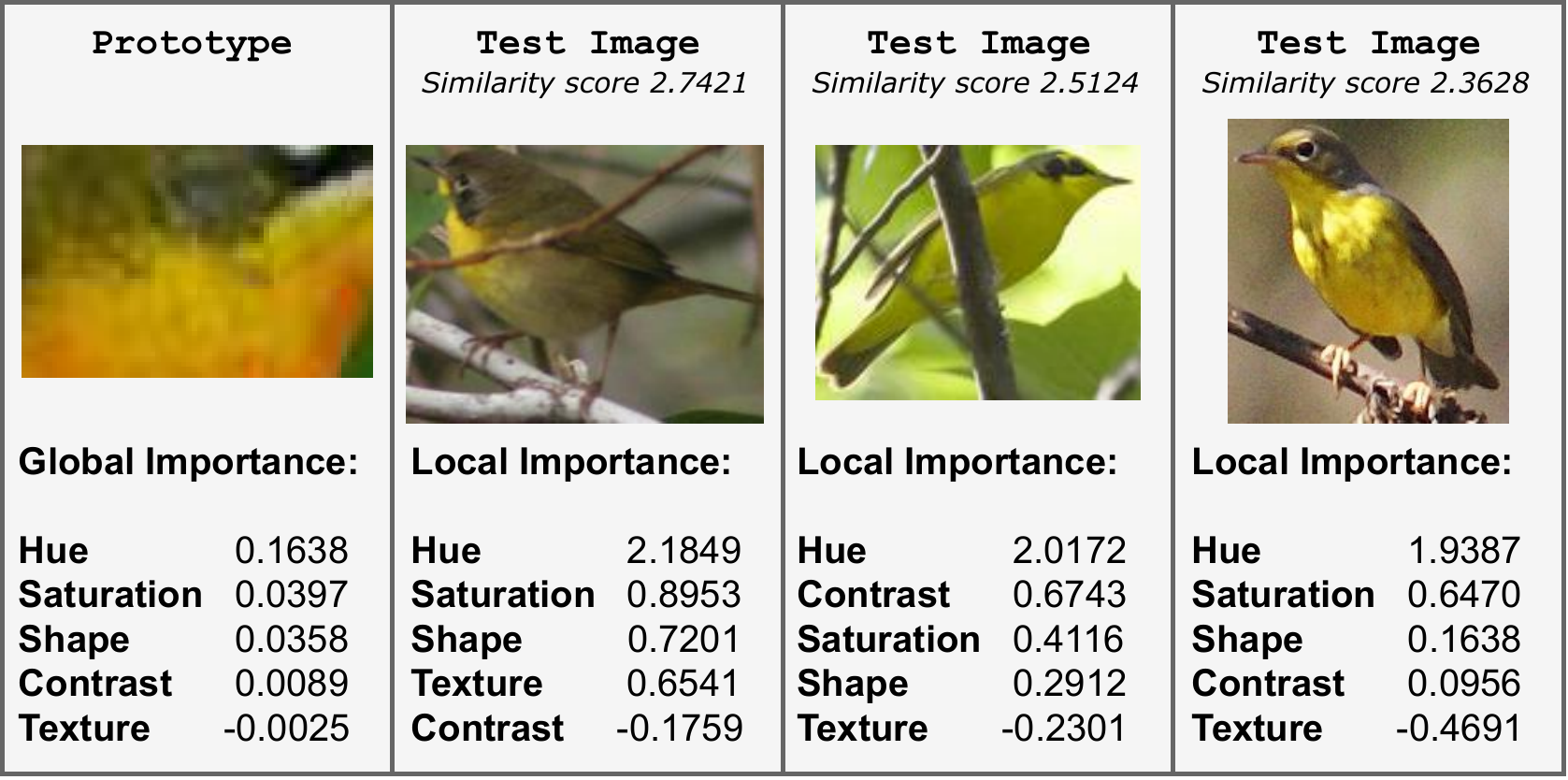}
\caption{As expected, the yellow hue is the dominant characteristic, and the local explanations correspond with the global importance.}
\label{fig:local_expl_yellow}
\end{subfigure}
\begin{subfigure}[b]{0.99\linewidth}
\centering
\includegraphics[width=\linewidth]{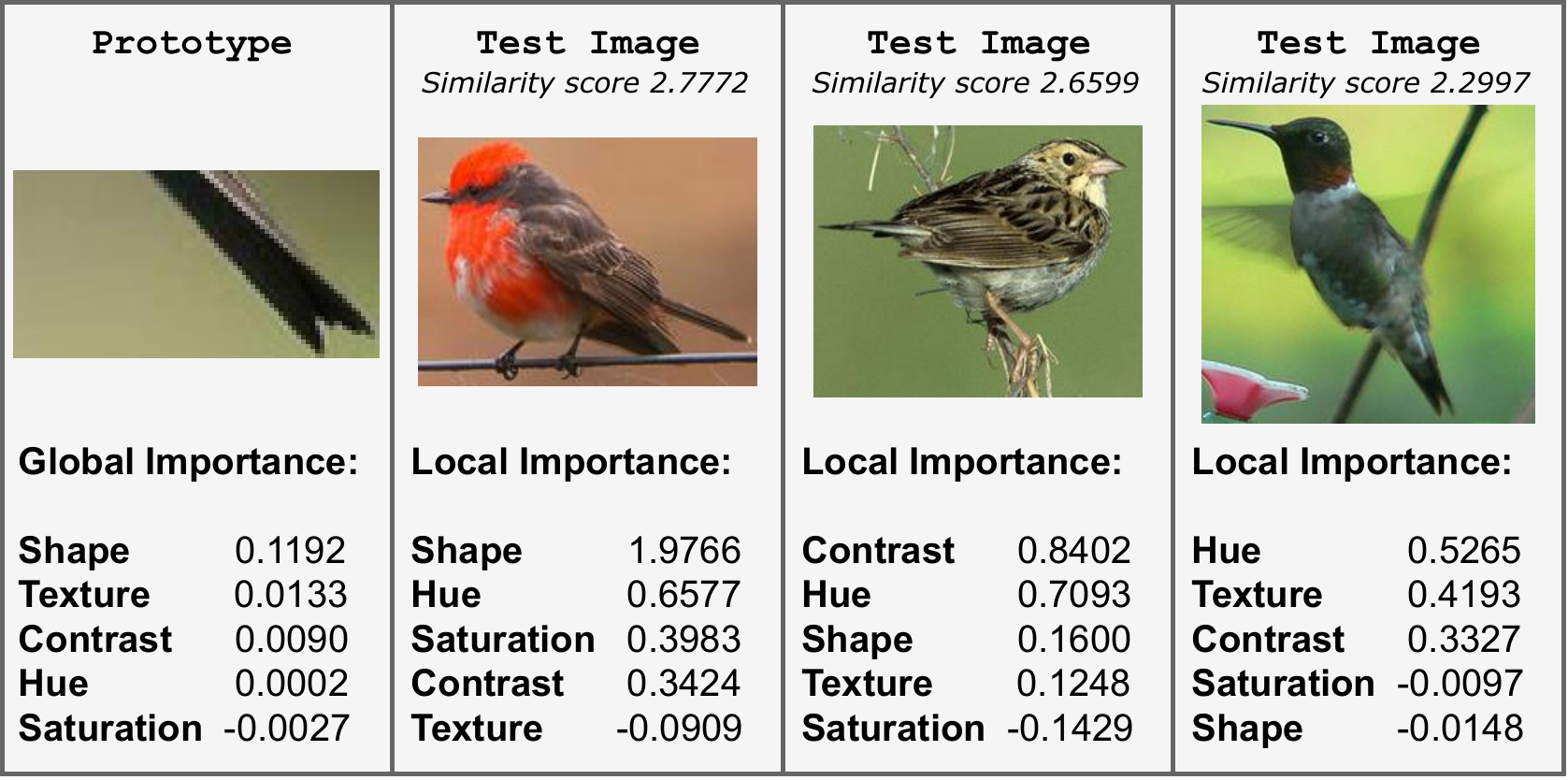}
\caption{Test images can get a high similarity score with the prototype for different reasons.}
\label{fig:diff_local_expl}
\end{subfigure}
\caption{Similarity between a class-specific prototype and test images from different classes explained by our local importance scores.}
\end{figure}

Our local explanations can also be useful to confirm the user's expectations, such as the importance of the yellow colour for the prototype in Figure~\ref{fig:local_expl_yellow}. The explanations are however especially insightful when the given similarity score is in contrast with human perceptual similarity and an explanation is needed. Figure~\ref{fig:diff_local_expl} explains why different test images received a high similarity score by ProtoPNet. The global importance scores of the prototype indicate that shape is most important for this prototype (cf. Sect.~\ref{sec:results_global_explanations}). However, some test images have a high similarity with the prototype for other reasons, such as hue, texture or contrast. Our approach therefore serves as an extension to a prototypical model, where the local importance scores explain a single prediction. 

\subsection{Analysing our Global Explanations}
\label{sec:results_global_explanations}
Since our trained ProtoPNet has 2000 prototypes, every test image will have 2000 local explanations for each of the five characteristics. Local explanations can therefore be useful to explain an unexpected result, but do not give a coherent, overall explanation of the model. Our methodology therefore also produces \emph{global} explanations that give an average view regarding the importance scores for each prototype. Specifically, the global importance scores are computed for each prototype by taking the weighted mean of all training images, as introduced in Section~\ref{sec:proto_anno}. Hence, these explanations can be generated \emph{before} applying the model to a test image, since global explanations are independent of test input. 

\subsubsection{Quantitative Evaluation}
To quantitatively evaluate our global explanations, we compute global importance scores twice: in addition to calculating the scores for $S_\text{train}$, the scores are for evaluation purposes also computed for $S_\text{test}$. 
We confirmed with a Shapiro-Wilk test that the global importance scores for each characteristic are normally distributed, such that we could apply the Welch t-test to confirm that there is no significant difference between the global importance scores of all prototypes calculated from $S_\text{train}$ and from $S_\text{test}$ for each characteristic (p-values $< 3.5^{-11}$). This verifies that the global importance scores are generalisable and robust, since the global explanation for a prototype does not significantly change when computed from a different image set. 

\begin{figure}
\centering
\includegraphics[width=0.475\textwidth]{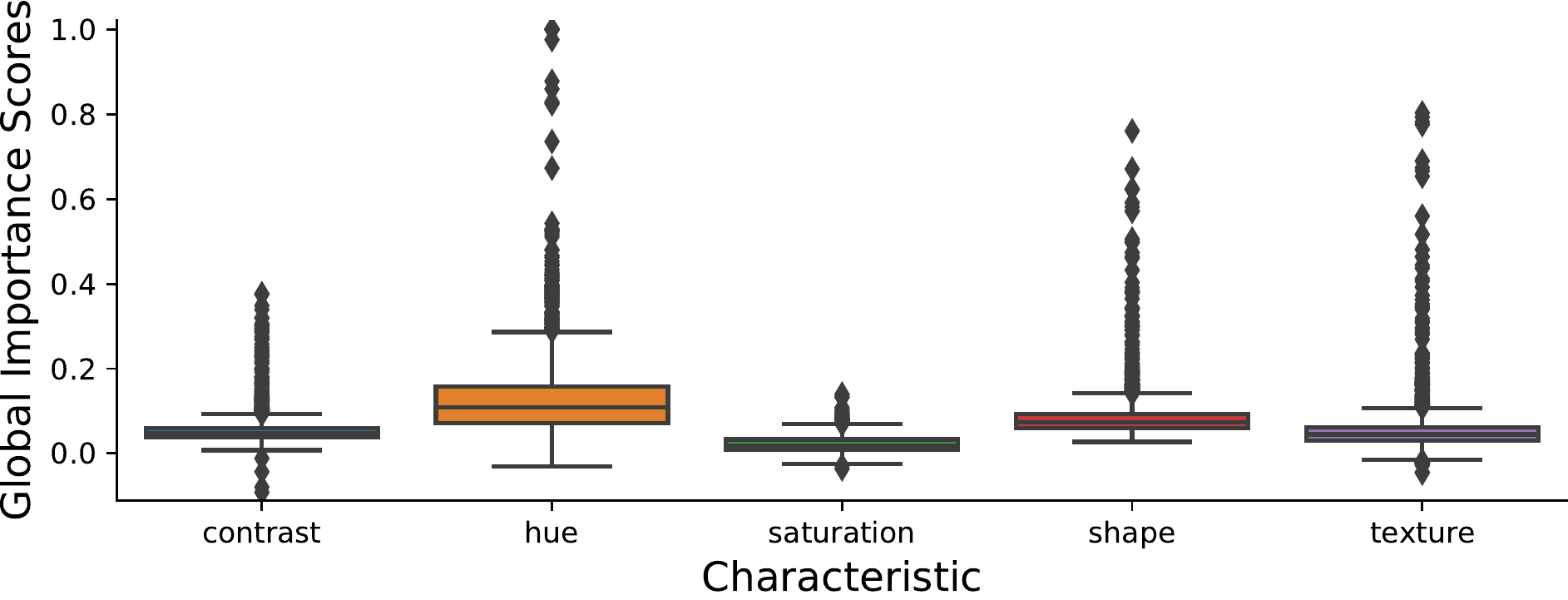}
\caption{Box plot of global importance scores across the training set for each visual characteristic.}
\label{fig:boxplotdiff}
\end{figure}

To get more insight as to how characteristics are used, we also analyse how the global importance scores are distributed per characteristic. 
Figure~\ref{fig:boxplotdiff} plots the distributions of global importance scores of all prototypes across the training set. It shows that the global scores are predominantly positive, which confirms our intuition that decreasing any of the visual characteristics usually leads to a lower similarity score.
It also shows that the mean and variability of importance scores is small for saturation and contrast, meaning that those characteristics only have a moderate influence on prototype similarity. In contrast, the high variability for shape, texture and especially hue means that these characteristics can be substantially important for some prototypes. This corresponds with the fact that hue and shape are considered more important and effective for humans in the data visualisation domain than saturation or contrast~\cite{munzner_visualization_2015}.

\subsubsection{Qualitative Evaluation}
\begin{figure}
\centering
\includegraphics[width=\linewidth]{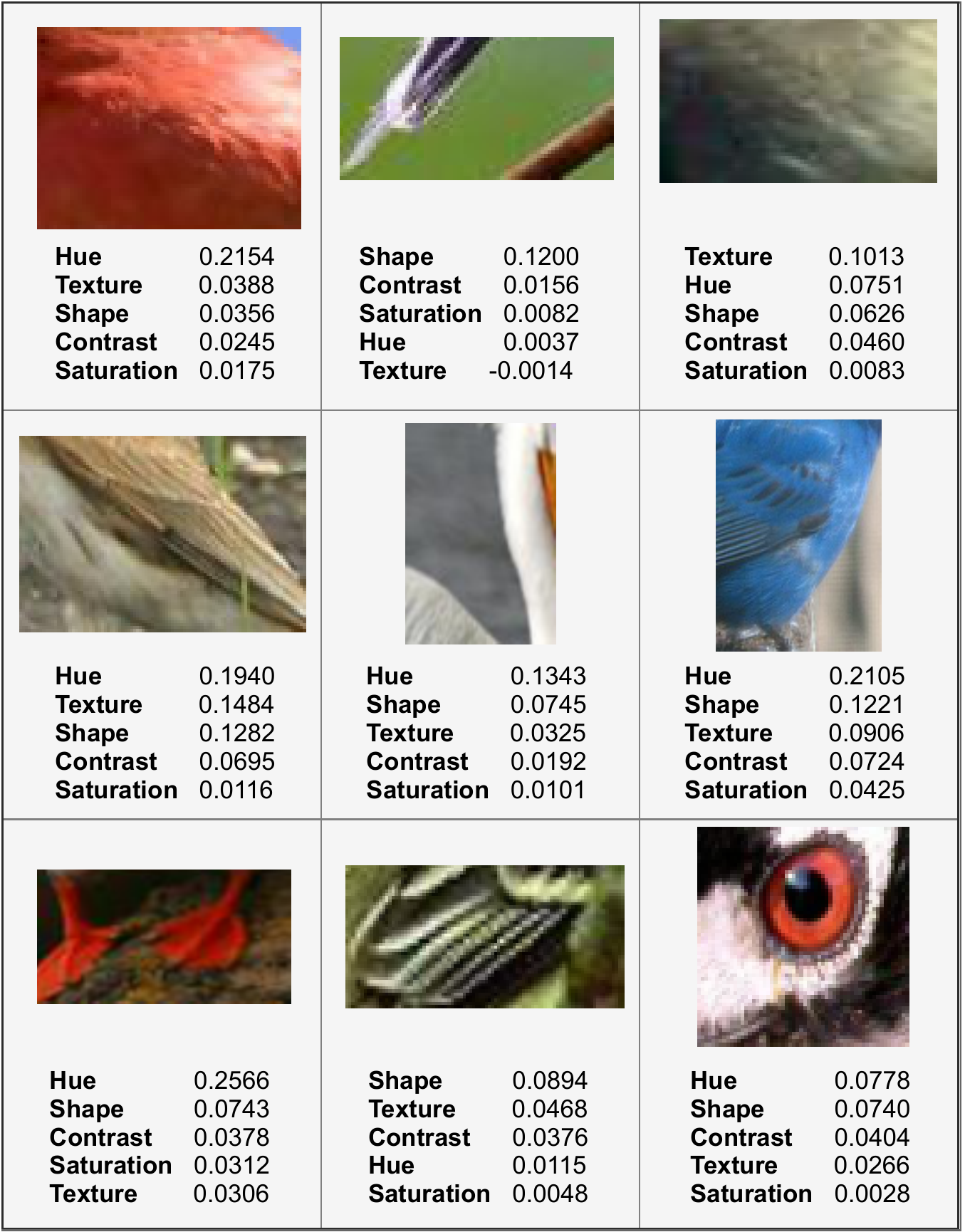}
\caption{Selection of prototypes explained with their global importance scores. Top row: Prototypes with predominantly a single important characteristic. Center row: Prototypes with intuitive explanations. Bottom row: ambiguous and potentially misleading prototypes.}
\label{fig:annotatedsample}
\end{figure}
\begin{figure}
\centering
\begin{subfigure}[t]{0.95\linewidth}
\includegraphics[width=\linewidth]{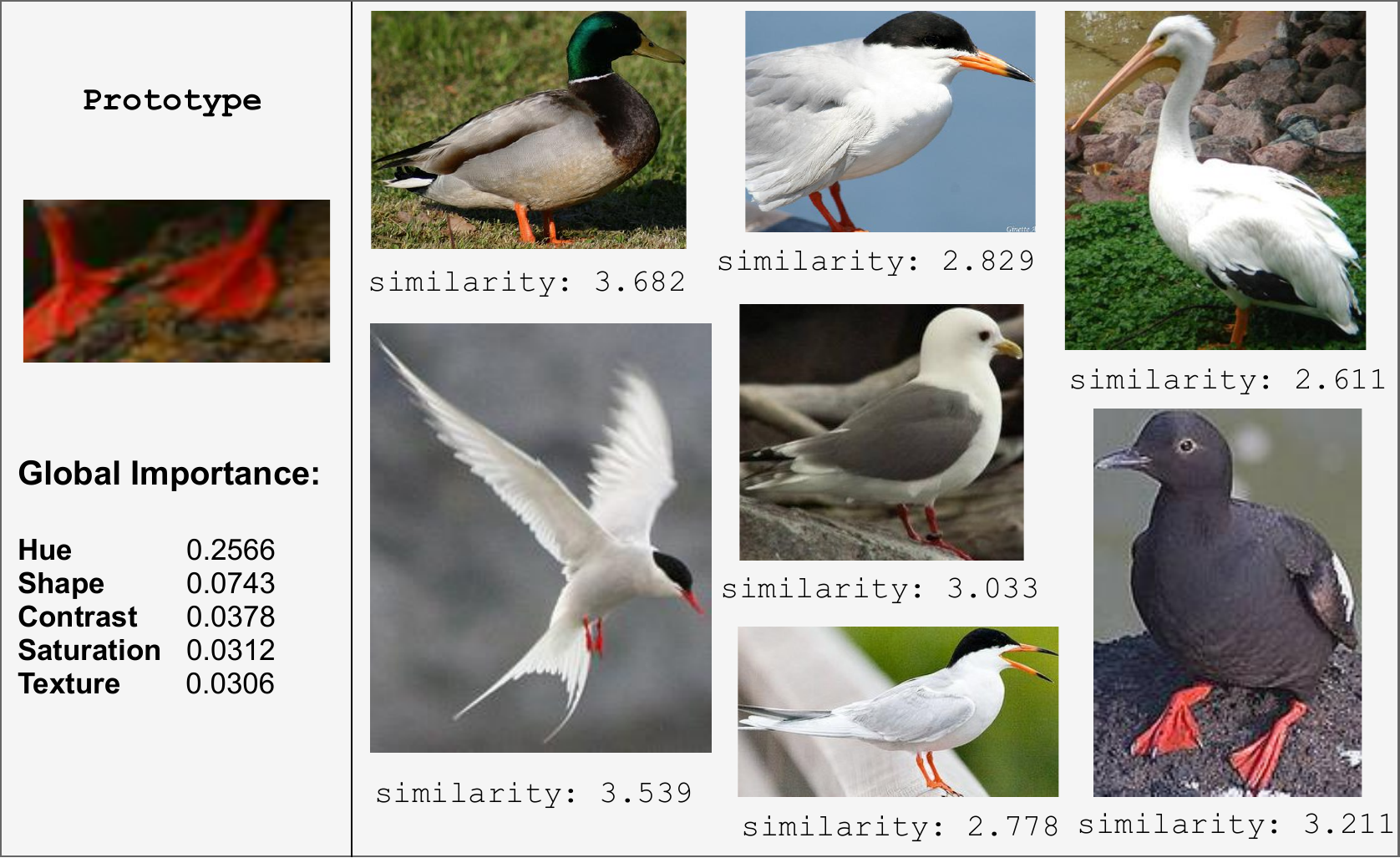}
\caption{The prototype indeed deems hue more important than shape.}
\label{fig:highest_redlegs}
\end{subfigure}
\begin{subfigure}[t]{\linewidth}
\centering
\includegraphics[width=0.95\linewidth]{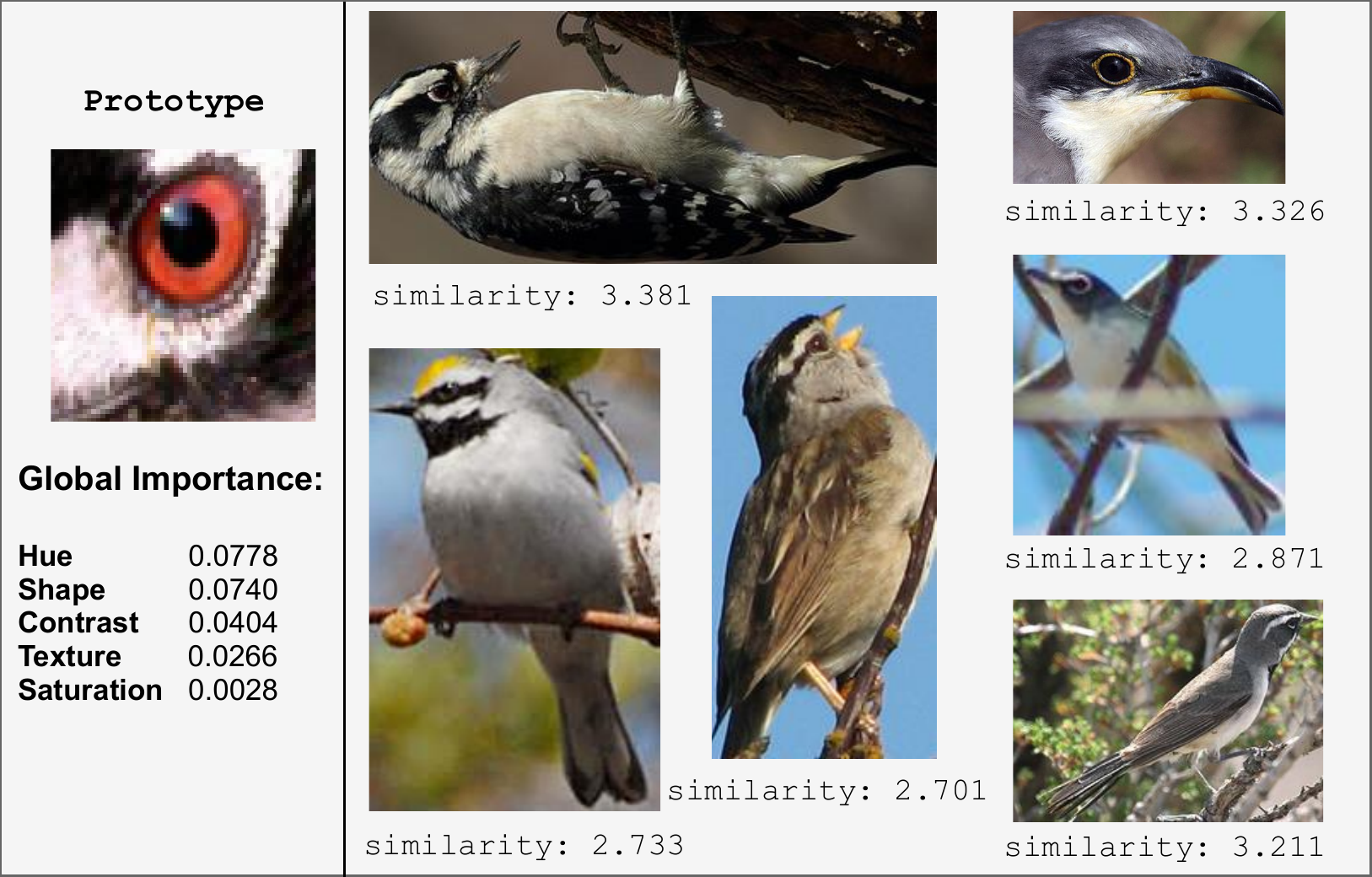}
\caption{The global score explains that the red hue from the eye is not that important, which is validated by near test images.}
\label{fig:highest_redeye}
\end{subfigure}
\caption{Test images from a different class than the prototype-class which have the highest similarity scores with the prototype.}
\end{figure}

Figure~\ref{fig:annotatedsample} shows a varied selection of prototypes with their global explanation. For the upper two rows, the importance of characteristics corresponds to the visually identifiable properties of the prototypes and hence, the explanations seem reasonable. However, the explanations in the bottom row might come as a surprise. A human might think that shape is important for the bottom left prototype and that the prototype resembles fin-footed birds. Our global explanation indicates that shape is of little importance and that colour hue is the dominant characteristic.
Since a ground-truth is not available, we verify the correctness of the global explanation by analysing test images that had a high similarity with the prototype. Although a prototype is trained to be class-specific, Figure~\ref{fig:highest_redlegs} shows images from the test set that are from a different class but still get assigned a high similarity score. These images confirm that the prototype deems hue important and therefore resembles \emph{red} feet, instead of webbed feet. 
The reverse is true for the bottom right prototype of Figure~\ref{fig:annotatedsample}. Humans could think that the prototype resembles a red eye, and would be surprised by a high similarity score with a black-eyed bird, and hence might lower their trust in the model. 
Our global importance scores indicate that the importance for hue is rather low, and hue and shape are of similar importance. When analysing birds that get assigned a high similarity with this prototype as shown in Figure~\ref{fig:highest_redeye},
it is easily verified that the prototype does indeed not represent a red eye, but dark and light stripes around an eye. 

These examples show that without our explanations, a user would not be aware of the meaning of a given prototype and correct \emph{simulatability}~\cite{Lipton2018mythos} would not be guaranteed. Our global explanations can therefore clarify visual prototypes and improve the simulatability of a prototype-based model. 

\subsection{Redundant Prototypes}
\label{sec:redundant_prototypes}
\begin{figure}
\centering
\includegraphics[width=\linewidth]{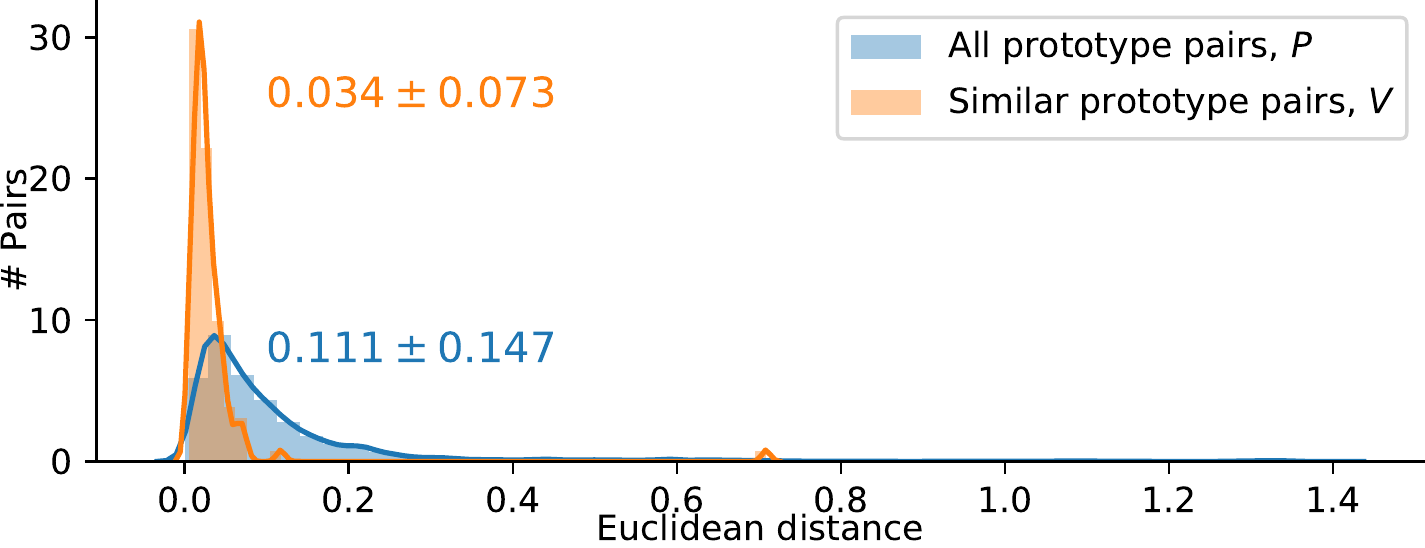}
\caption{Histogram with the distribution of Euclidean distances between the global explanations of two prototypes of the same class (a `pair'). }
\label{fig:redundancy_distribution}
\end{figure}

An interesting question is whether prototypes that are slightly different, deem the same visual characteristics important. If their global importance scores are different, then these prototypes complement each other, whereas similar explanations indicate prototype redundancy. Such prototype redundancy was also discussed by~\cite{10.1145/3292500.3330908}, and was found to increase with the number of prototypes per class. 
We consider prototypes to be visually similar when they are close to each other in the latent space learned by ProtoPNet. 
We measure the Euclidean distance between the latent representation of two prototypes of the same class (a `pair'). This gives $\binom{10}{2} = 45$ unique pairs per class, and $45 \cdot 200 = 9000$ pairs in total. Let $P$ be the set of unique pairs of two prototypes from the same class, such that $|P| =9000$, and $V \subset P$ the set of pairs with two visually similar prototypes. We consider a pair of two prototypes $i$ and $j$ \emph{identical} when the Euclidean distance in latent space $d_{i,j} = 0$ and \emph{visually similar but not identical} when $d_{i,j} < \tau$ and $d_{i,j} > 0$, where $\tau = 0.15$ is found to be a suitable threshold for perceptual similarity. This gives 63 pairs of identical prototypes and $|V| = 93$ unique pairs of 164 visually similar prototypes. 
To evaluate whether these visually similar prototypes also have similar global explanations, we consider the global importance scores of a prototype as a vector of length 5 and calculate the Euclidean distance between the global explanations of two prototypes. 
The orange plot in Figure~\ref{fig:redundancy_distribution} shows that most pairs with visually similar prototypes have a small distance between their global importance scores, which is not the case in general (blue). Therefore, these prototypes might be redundant and unnecessarily increase explanation size. Additionally, a few pairs in the orange plot have dissimilar explanations (distance of roughly 0.7) and therefore complement each other. 

\section{Conclusion and Future Work}
A prototype-based image recognition model learns visual prototypes and localises a patch in a test image that looks alike a prototype to assign it a similarity score. We argue that these prototypes should be explained with respect to the model's reasoning and extend localisation with explanation. We presented an automated approach to explain visual prototypes learned by any prototypical image recognition model. Our method automatically modifies the hue, texture, shape, contrast or saturation of an image, and evaluates the model's similarity score with a prototype. In this way, we identify which visual characteristics of a prototype the model deems important. 
We applied our method to the prototypes learned by the Prototypical Part Network (ProtoPNet)~\cite{chen2019looks}.
The importance of visual characteristics identified by our explanations often corresponded to the visually perceptible properties of the prototypes, showing that our explanations are reasonable. 
We also showed that perceptual similarity for humans can be different from the similarity learned by the model, indicating the need for explaining the model's reasoning. 
Such `misleading' prototypes will hinder correct simulatability and only visualising prototypes can be insufficient for understanding why the model considered a prototype and an image highly similar. To the best of our knowledge, we are the first to address such ambiguity of visual prototypes and the elegant simplicity of our approach makes it a suitable stand-alone solution.

A limitation of our method is the extra computation: each image needs to be modified for each of the five visual characteristics, and this modified image should be forwarded through the prototypical model to compare similarity scores. However, we think the extra computational complexity is justifiable given the extra insights our method provides. Furthermore, because of the stand-alone nature of our method, it can be applied to any prototypical image recognition method, including ProtoPNet~\cite{chen2019looks} and ProtoTree~\cite{nauta2020neural}. 
Our approach can also easily be extended with more visual characteristics or other image modifications a user is interested in.

Future work concerns the potential interactions between characteristics. Our importance scores currently assume that characteristics from image modifications are mutually exclusive. However, denoising the image to lower its texture could also slightly influence shape. We implemented the image modifications in such a way to limit interactions between characteristics as much as possible, but future analysis could determine to what extent visual characteristics are correlated.

\newpage
\bibliographystyle{ACM-Reference-Format}
\bibliography{bibliography.bib}
\end{document}